\documentclass[letterpaper]{article}
\usepackage{aaai17}
\usepackage{times}
\usepackage{helvet}
\usepackage{courier}
\RequirePackage{snapshot}

\usepackage{subcaption}

\usepackage{moreverb}
\usepackage{verbatim}

\usepackage{acronym}
\usepackage{enumitem}
\usepackage{color}

\usepackage{latexsym}
\usepackage{tabularx}
\usepackage[sharp]{easylist}
\usepackage{etoolbox}

\usepackage{siunitx}
\usepackage[usenames,dvipsnames,table]{xcolor}

\usepackage{amsmath}
\usepackage{amsfonts}
\usepackage{amssymb}
\usepackage{amsthm}

\usepackage{algorithmic}
\usepackage{algorithm}
\usepackage{cases}

\usepackage[colorinlistoftodos,disable]{todonotes}

\newcommand{\cd}[1]{\texttt{#1}}

\usepackage{styles/mathdefs}

\renewcommand{\eqref}[1]{(\ref{eq:#1})}
\newcommand{\secref}[1]{\S\ref{sec:#1}}

\newcommand{\figref}[1]{Fig.~\ref{fig:#1}}

\let\OldEasylist\easylist
\let\OldEndEasylist\endeasylist
\renewenvironment{easylist}{%
    \OldEasylist%
    \ListProperties(Style**=\color{blue},Start1=1)%
}{%
    \OldEndEasylist%
}%

\begin{document}



\title{Semi-Automated Annotation of Discrete States\\in Large Video Datasets}
\author{
  Lex Fridman\\
  Massachusetts Institute of Technology\\
  \texttt{fridman@mit.edu}
  \And
  Bryan Reimer\\
  Massachusetts Institute of Technology\\
  \texttt{reimer@mit.edu}
}


\maketitle

\begin{abstract}
  We propose a framework for semi-automated annotation of video frames where the video is of an object that at any point in time can be labeled as being in one of a finite number of discrete states. A Hidden Markov Model (HMM) is used to model (1) the behavior of the underlying object and (2) the noisy observation of its state through an image processing algorithm. The key insight of this approach is that the annotation of frame-by-frame video can be reduced from a problem of labeling every single image to a problem of detecting a transition between states of the underlying objected being recording on video. The performance of the framework is evaluated on a driver gaze classification dataset composed of 16,000,000 images that were fully annotated through 6,000 hours of direct manual annotation labor. On this dataset, we achieve a 13x reduction in manual annotation for an average accuracy of 99.1\% and a 84x reduction for an average accuracy of 91.2\%.
\end{abstract}


\section{Introduction and Related Work}\label{sec:introduction}

The biggest of ``big data'' is video \cite{mayer2013big}. 300 hours of video are uploaded to YouTube every minute and,
in 2014, YouTube and Netflix accounted for 50\% of all peak period downstream traffic in North America
\cite{liu2015quantitative}. Like most big data, video is largely unstructured, unlabeled, and unprocessed. However, the
proven effectiveness of supervised machine learning in image processing in the last 15 years has shown promise that
computer vision can help automate the process of interpreting the content of big video data \cite{jordan2015machine}.

As the availability of large-scale video datasets has become widespread over the past decade, semi-automated annotation
of images and videos has received considerable attention in computer vision literature
\cite{yuen2009labelme,kavasidis2012semi}. In the image domain, the focus has been on segmenting the
image into distinct entities and assigning multiple semantic labels to each entity
\cite{ivasic2015knowledge}. In the video domain, the focus has been on labeling segmented entities in
select keyframe images from the video and propagating the annotation in the keyframes to the frames in-between via
linear interpolation, tracking, or time-based regularization \cite{bianco2015interactive}. Multi-label annotation and
tracking are ambitious and complex efforts that are essential for a general video interpretation framework. Our paper
focuses on a narrow but important subset of video data: where a single object is being recorded and that object can be
labeled as belonging to one of a finite number of discrete states. We propose a novel framework for semi-automated
annotation of that type of video. The two key aspects of our approach is: (1) we reduce the problem of video annotation
into change detection and (2) we form a Hidden Markov Model (HMM) for the changes between the states in order to predict
when a sequence of classifier decisions correspond to a sequence of stable state self-transitions.

The proposed semi-automated video annotation framework is evaluated on 16,000,000 video-frames of driver faces collected
over 150 hours of field driving. The classification of driver gaze regions is an area of increasing relevance in the
pursuit of accident reduction. The allocation of visual attention away from the road has been linked to accident risk
\cite{victor2014analysis} and a drop in situational awareness as uncertainty in the environment increases
\cite{senders1967attentional}.

The problem of gaze tracking from monocular video has been investigated extensively across many domains
\cite{gaur2014survey,sireesha2013survey}. We chose driver gaze classification as the case study for the proposed
framework for two reasons. First, gaze tracking from video in the driving context is a difficult problem due especially
to rapidly varying lighting conditions. Other challenges, common to other domains, include unpredictability of the
environment, presence of eyeglasses or sunglasses occluding the eye, partial occlusion of the pupil due to squinting,
vehicle vibration, image blur, poor video resolution, etc. We consider the challenging case of uncalibrated monocular
video because it has been and continues to be the most commonly available form of video in driving datasets due to low
equipment and installation costs. Second, we have 150 hours of double annotated and mediated data. This allows us to
evaluate the key trade-off of the semi-automated annotation problem: the reduction in human effort and the
classification accuracy achievable under that reduction. The tradeoff curve we present at the end of the paper shows a
13x reduction in human effort for an average accuracy of 99.1\% and a 84x reduction for an average accuracy of
91.2\%. This improves significantly on the best result from a similar domain of head pose estimation where 3.3x
reduction of manual effort and an accuracy of 97\% was achieved on a set 108,000 images \cite{demirkus2014robust}. The
strengths and drawbacks of the proposed approach can be summarized as follows:

\paragraph{Strengths}

\begin{enumerate}
\item \textbf{Novel approach to video annotation:} To the best of our knowledge, ours is the first semi-automated
  discrete state annotation method in video. The key contribution of our work is this idea itself: simplify the problem
  of video annotation into two problems: (1) discrete object state classification and (2) change detection. In many
  computer vision application domains these two individual problems have well-developed robust, accurate solutions.
\item \textbf{Novel approach to gaze classification:} A lot of work has been done on gaze estimation and classification,
  but to the best of our knowledge, temporal estimation of gaze regions as a set of discrete states with detectable
  state transitions has not been done before. That's why gaze classification was the chosen case study. We believe that
  many other video analysis applications are similar to it and could be improved with this approach.
\item \textbf{State-of-the-art performance:} With the change detection method, to the best of our knowledge, we achieve the best
  gaze classification results on any large in-the-wild datasets to date.
\item \textbf{Dataset size:} The 16,000,000 image evaluation dataset is, by 2 orders of magnitude, the largest annotated gaze
  dataset we are aware of. We seek to show that this provides a vision for large ever-growing datasets where supervised
  learning can truly shine.
\end{enumerate}

\paragraph{Drawbacks}

\begin{enumerate}
\item \textbf{Theoretical bounds:} There are no theoretical guarantees on the performance of the overall framework. This
  is a fundamental shortcoming of an approach  that uses supervised learning methods. It relies on the ability of the two
  underlying classification algorithms to generalize sufficiently well over the data to be of assistance to the human
  annotator on labeling future data. We will make clear in the paper that the framework makes no guarantees on
  performance.
\item \textbf{Parameters:} There are several parameters in the framework with no automated way of tuning those
  parameters for a specific domain. This is a big drawback of many learning approaches for real-world datasets. In our
  empirical evaluation (as shown in \figref{annotation-performance}), the choices of parameters never resulted in
  significantly sub-Pareto-optimal performance, but same as in drawback \#1, no provable guarantees can be provided.
\end{enumerate}


\section{Semi-Automated Annotation Framework}\label{sec:framework}

\begin{figure}
  \centering
  \includegraphics[width=\columnwidth]{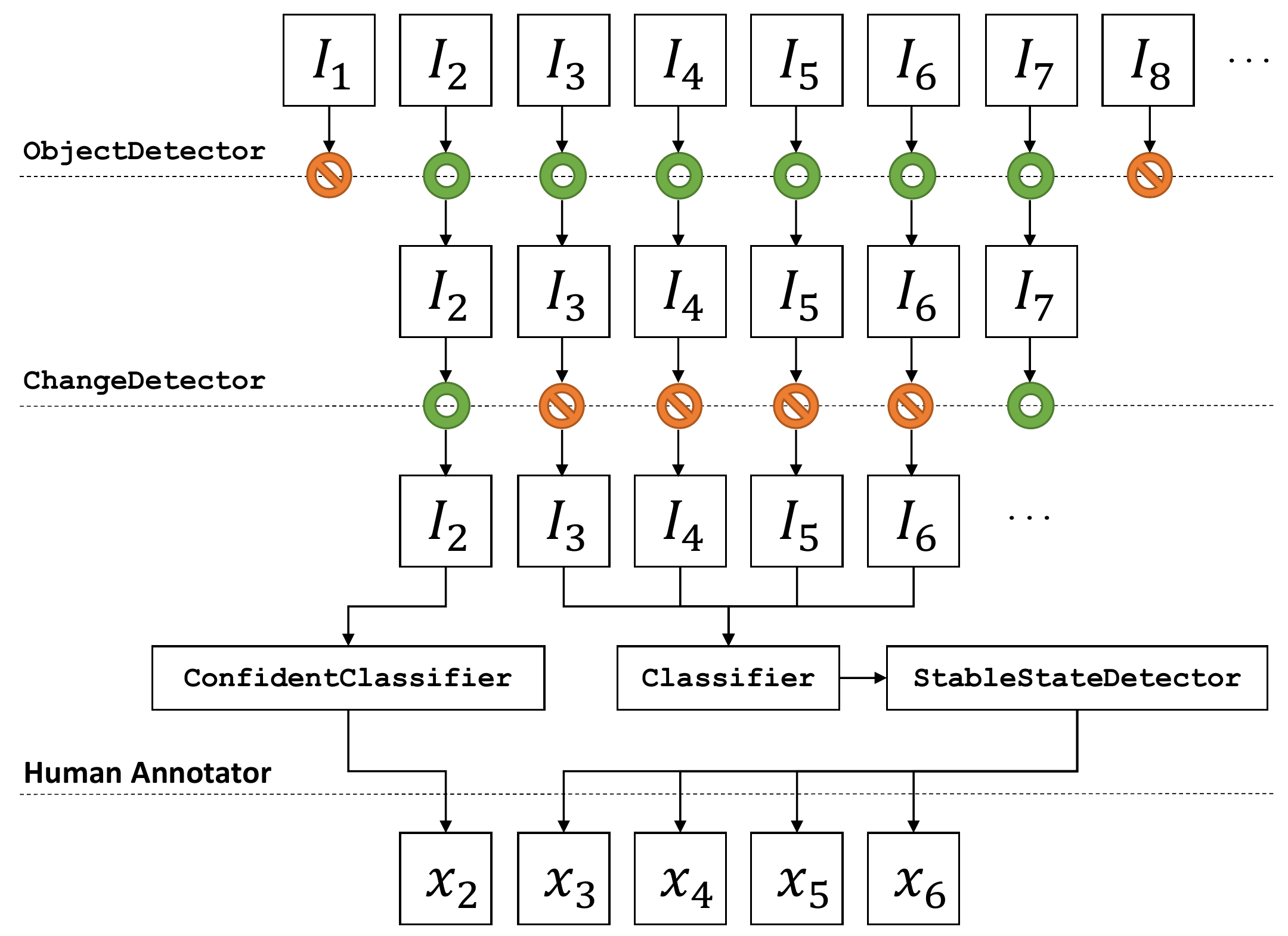}
  \caption{Illustrative diagram of the semi-automated annotation pipeline. The input of the pipeline are the video
    frames at the top and the output of the pipeline is the per-frame classification of object state at the bottom. The
    \cd{ObjectDectector} and \cd{ChangeDetector} produce binary $\{0, 1\}$ outputs which are indicated in the diagram as
    crossed-circles for $0$ and empty circles for $1$.}
  \label{fig:diagram}
\end{figure}

\subsection{State Model and Stable State Detection}\label{sec:state-model}

The video to be annotated is assumed to be of an object that can be labeled at any moment in time as in one instance of
state space $S$. We formulate the dynamics of this discrete state system and its annotation as a Hidden Markov Model
(HMM) \cite{rabiner1986introduction}. The hidden state is the correct classification of every video frame. The observed
output is the estimated class of the image according to the intermediate steps of the pipeline described in
\secref{pipeline}.

During the annotation process, we continually update initial probabilities $\pi_i$ of being in state $i$ and transition
probabilities $a_{i,j}$ of transitioning from state $i$ to state $j$. Given the observed classifier decisions $y_1,
\dots, y_m$, the most likely state sequence $x_1, \dots, x_m$ that produces the observed decisions is given by:

\begin{equation}\label{eq:viterbi}
  \begin{array}{rcl}
    V_{1,k} &=& \mathrm{P}\big( y_1 \ | \ k \big) \cdot \pi_k \\
    V_{t,k} &=& \max_{x \in S} \left(  \mathrm{P}\big( y_t \ | \ k \big) \cdot a_{x,k} \cdot V_{t-1,x}\right)
  \end{array}
\end{equation}

\noindent where $V_{t,k}$ is the probability of the most probable state sequence responsible for the first $t$
observations that have $k$ as its final state. This recurrence relation is used to compute the Viterbi path
\cite{shinghal1980sensitivity}, the most likely sequence of states that results in the observed classifier decisions. We
denote the probability of the Viterbi path as $V^*$.

The probability that a set of observations is associated with a sequence of unchanged hidden states, assuming the first
and state is $k$ is given by:

\begin{equation}\label{eq:unchanged}
  \begin{array}{rcl}
    V_{1,k} &=& \mathrm{P}\big( y_1 \ | \ k \big)\\
    V_{t,k} &=& \mathrm{P}\big( y_t \ | \ k \big) \cdot a_{k,k} \cdot V_{t-1,k}
  \end{array}
\end{equation}

The probability of an unchanged sequence of hidden states $k$ normalized by $V^*$ is denoted by $V^u$. This value is
used in \secref{pipeline} to determine whether to pass the set of images associated with the sequence of observed
classifier decision to the human annotator.

\subsection{Semi-Automated Annotation Pipeline}\label{sec:pipeline}

Given a sequences of temporally-adjacent images $I_t$ for $t\in{1, \dots, T}$, the annotation pipeline detail in this
section and illustrated in \figref{diagram} produces hidden state predictions $x_t$ for $t\in{1, \dots, T}$. We first
describe each of the five component algorithms, and then specify how they are incorporated in the annotation framework.

\begin{itemize}
\item \cd{ObjectDetector}: $I_t \rightarrow \{0,1\}$\\
  Given a single image, this algorithm provides a binary decision on whether the object of interest is present in the
  image (value of 1) or is absent in the image (value of 0). In the case of gaze classification, this algorithm is a
  face detector.
\item \cd{Classifier}: $I_t \rightarrow S$\\
  Given a single image, this algorithm provides a classification of what hidden state $i \in S$ the object in the image is in.
\item \cd{ConfidentClassifier}: $I_t, c_{\min} \rightarrow S$ \\ Given a single image and a confidence threshold
  $c_{\min}$, this algorithm provides a classification of what hidden state $i \in S$ the object in the image is in if
  the classification is above a confidence threshold. The purpose of this algorithm is to replace the human in cases
  when a confident classification decision can be made.
\item \cd{ChangeDetector}: $I_{t-1}, I_t, \delta_{\min} \rightarrow \{0, 1\}$
  Give an image at time $t$ and an image immediately before it, this algorithm provides a binary decision on whether a change
  of state is detected in the image at time $t$. The change detector uses the threshold $\delta_{\min}$ to convert the
  continuous change detector output to a binary decision.
\item \cd{StableStateDetector}:\\$\{I_1, I_2, ..., I_T\}, V^u_{\min} \rightarrow \{0, 1\}$\\
  Given a temporally-adjacent sequence of images and a likelihood threshold $V^u_{\min}$, this algorithm makes a
  decision on whether the sequences of images correspond to a sequence on unchanged hidden states (see
  \secref{state-model}).
\end{itemize}

The three parameters which control the behavior of the framework are the change detector threshold $\delta_{\min}$, the
classifier confidence threshold $c_{\min}$, and the stable state detector threshold $V^u_{\min}$. Changing these values
results in various points along the tradeoff in human effort and classification effort show in the parametric
plot in \figref{annotation-performance}.

As illustrated in \figref{diagram}, the above specified algorithms are incorporated in the following data flow:

\begin{enumerate}
\item \textbf{Sequence Segmentation:} The video data is segmented into continuous sequences of images where the
  \cd{ObjectDetector} successfully detects the object of interest.
\item \textbf{Change Detection and Classification:} Detect the changes in the image sequences using \cd{ChangeDetector} and the threshold
  $c_{\min}$. Use \cd{ConfidentClassifier} to classify each of the 2 images before and after the detected change. If no confident
  classification can be made, these 5 images (2 prior, 1 current, and 2 subsequent images) are sent to the manual annotation queue.
\item \textbf{Change Verification:} If two adjacent change-points do not share a connecting state, all of the images
  between the two change-points are sent to the manual annotation queue.
\item \textbf{Stable State Detection:} Classify each state between the detected changes using the \cd{Classifier}. These
  classification decisions are the observations of the HMM described in \secref{state-model}. If the normalized
  probability $V^u$ of an unchanged state is above the threshold $V^u_{\min}$, then the states between the two detected
  changes are labeled according to the estimated hidden state.
\item \textbf{Human Annotation:} Any images added to the manual annotation queue in the previous 4 steps are double
  annotated and mediated by a human.
\end{enumerate}

The performance of the overall system depends on the accuracy of each of the five algorithms defined above. As the
annotation process proceeds, the manually annotated images should be frequently used to re-train each of the algorithms.

\section{Case Study: Driver Gaze Classification}\label{sec:driver-gaze}

\subsection{Dataset}\label{sec:dataset}

Evaluation of the semi-automated annotation framework is carried out on a dataset of 150 hours of video data spanning
244 different drivers traversing over 10,000 miles of highway. For each subject, the collection of data was carried out
in one of a set of study vehicles: 2013 Chevrolet Equinox, 2013 Volvo XC60, 2014 Chevy Impala, or 2014 Mercedes CLA
(randomly assigned).  The data was double manually annotated of driver glances transitions during secondary task periods
(at a resolution of sub-200ms) into one of 11 classes (road, center stack, instrument cluster, rearview mirror, left,
right, left blindspot, right blindspot, passenger, uncodable, and other). Any discrepancies between the two annotators
were mediated by an arbitrator.

\subsection{Gaze Classification and Change Detection}

\begin{figure}
  \centering
  \includegraphics[width=1.5in]{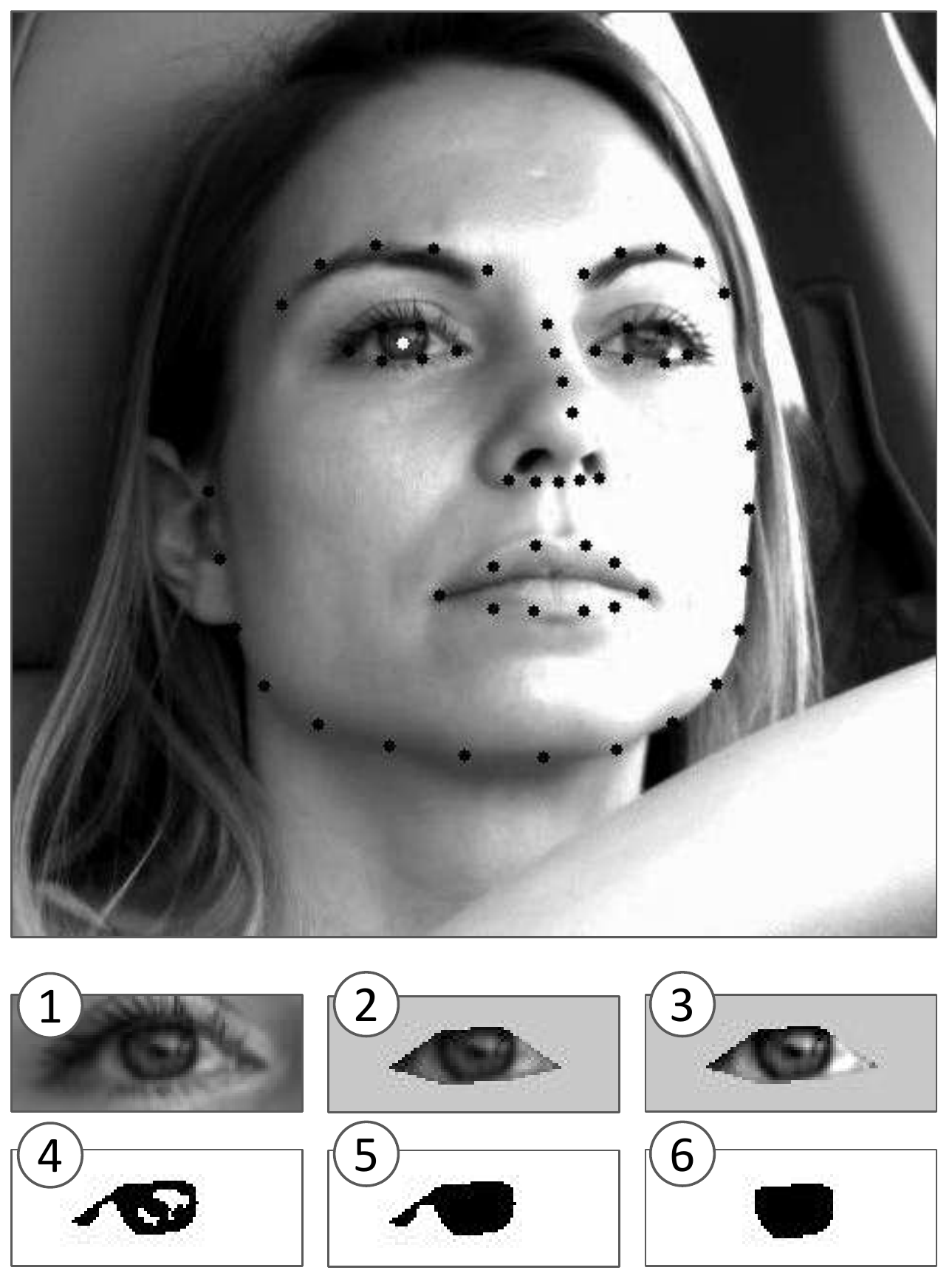}
  \caption{Example of facial landmark alignment and the intermediate steps of the pupil
    detection. The black dots designate the landmarks and the single white dot designates the pupil position in the
    right eye. Below the image of the face image are 6 steps of the pupil detection.}
  \label{fig:pupil}
\end{figure}

\newcommand{\changewidth}{0.95\columnwidth}
\begin{figure*}
  \centering
  \begin{subfigure}[t]{\changewidth}
    \includegraphics[width=\textwidth]{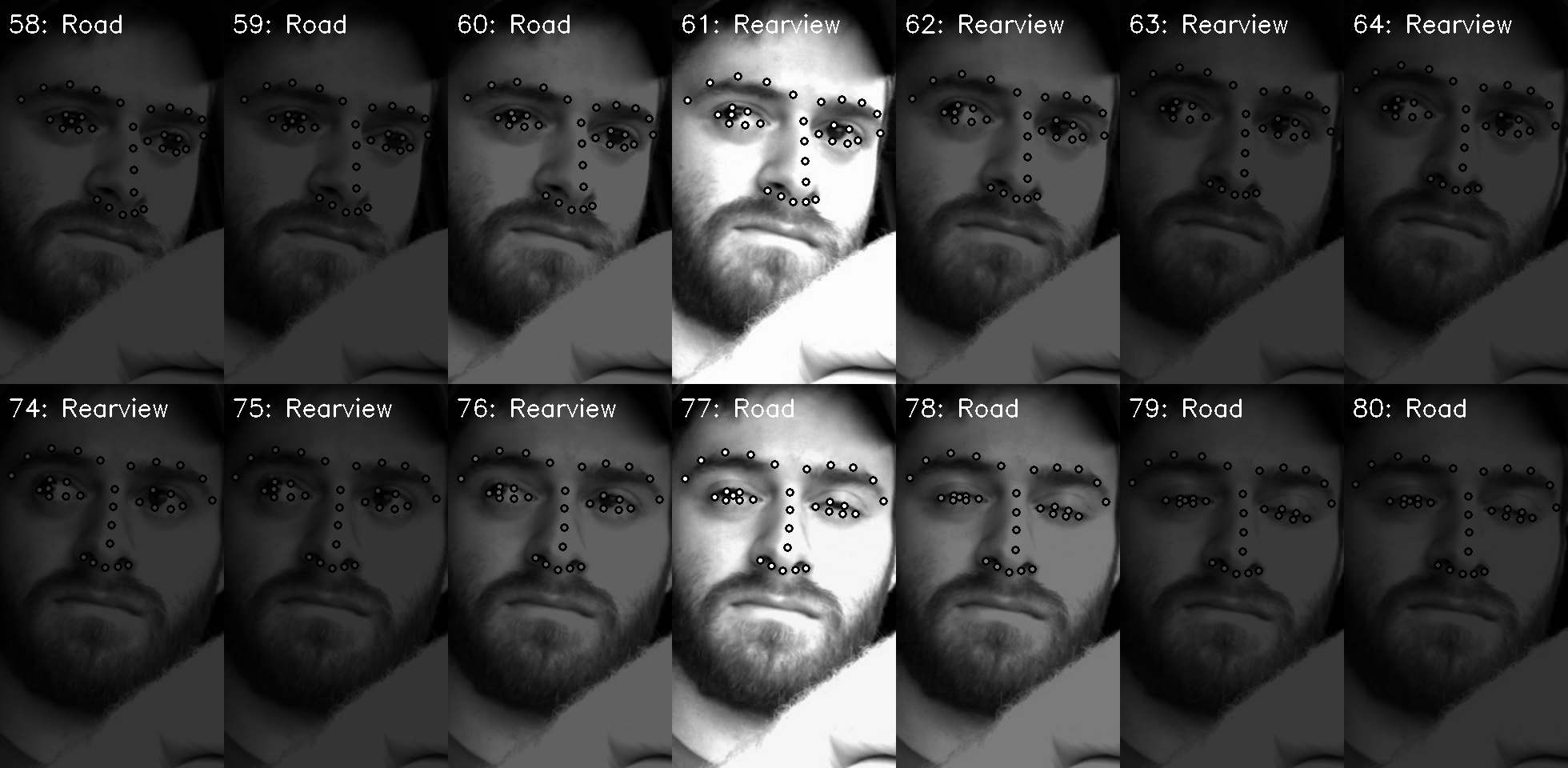}
    \caption{Two sequences of images showing a state transition (1) from ``Road'' to ``Rearview Mirror'' and (2) from
      ``Rearview Mirror'' to ``Road''. The transparency of the image (over a black background) is proportional to the
      value produced by the change detection classifier. The bright image in the middle of each sequence is the one
      predicted to be where the state transition occured.}
  \end{subfigure}\hspace{0.2in}
  \begin{subfigure}[t]{\changewidth}
    \includegraphics[width=\textwidth]{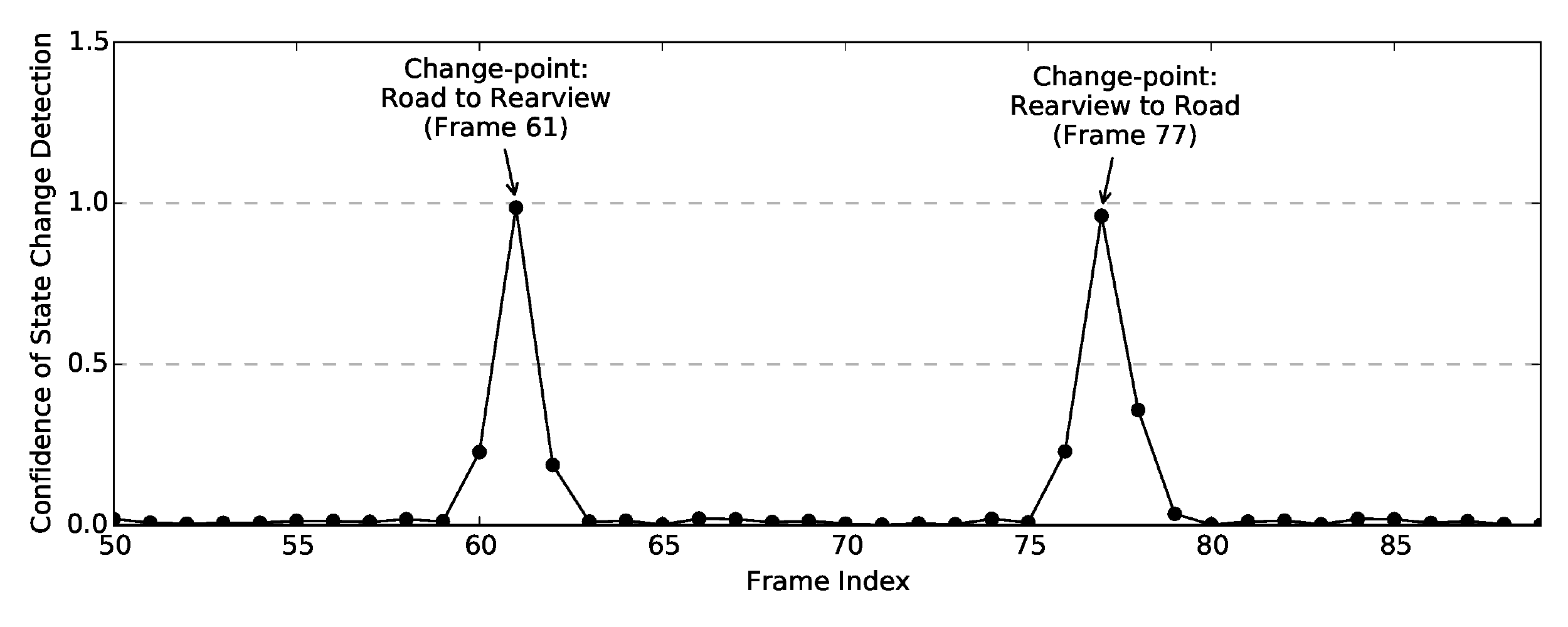}
    \caption{The output of the change detection classifier where a value of 1 means a change-point was detected and a
      value of 0 means a no change-point was detected. The x-axis of this plot corresponds to the frame indecies marked
      in each image above the plot.}
  \end{subfigure}
  \caption{Representative example of the change detection algorithm output on two sequences of images.}
  \label{fig:change-detection}
\end{figure*}


The application of the framework described in \secref{framework} requires the definition of \cd{ObjectDetector},
\cd{Classifier}, \cd{ConfidentClassifier}, \cd{ChangeDetector}, and \cd{StableStateDetector} specific to the gaze
classification problem. We consider the problem of classifying driver gaze into six regions: Road, Center Stack,
Instrument Cluster, Rearview Mirror, Left, and Right. These six region forms the state space $S$ for the HMM in the
implementation of the \cd{StableStateDetector} algorithm.

\subsubsection{Face Detection}

For the gaze classification problem, the object is the human face, and so the \cd{ObjectDetector} algorithm is a face
detector. The environment inside the car is relatively controlled in that the camera position is fixed and the driver
torso moves in a fairly contained space. Thus, a camera can be positioned such that the driver's face is always fully or
almost fully in the field of view. The face
detector in our pipeline uses a Histogram of Oriented Gradients (HOG) combined with a linear SVM classifier.

\subsubsection{Feature Extraction}\label{sec:feature-extraction}

The steps in the gaze region classification pipeline are: (1) face alignment, (2) pupil detection, (3) feature
extraction and normalization, and (4) classification. The face image in \figref{pupil} is an example of the result
achieved in the first three steps of the pipeline: going from a raw video frame with a detected face to extracted face
features and pupil position. The relative orientation of facial features serves as a proxy for ``head pose'' and the
relative orientation of pupil position serves as a proxy for ``eye pose''. We discuss each of the six steps in the
pipeline in the following sections.

Face alignment in our pipeline is performed on a 68-point Multi-PIE facial landmark mark-up used in the iBUG 300-W
dataset \cite{sagonas2013300}. These landmarks include parts of the nose, upper edge of the eyebrows, outer and inner
lips, jawline, and parts in and around the eye. The selected landmarks are shown as black dots in \figref{pupil}. The
algorithm for aligning the 68-point shape to the image data uses a cascade of regressors as described in
\cite{kazemi2014one} and implemented in \cite{dlib09}.

The problem of accurate pupil detection is more difficult than the problem of accurate face alignment, but both are not
always robust to poor lighting conditions. Therefore, the secondary task of pupil detection is to flag errors in the
face alignment step that preceded it. The face is detected in 79.4\% video frames but only 61.6\% of the original frames
pass the pupil detection step.

We use a CDF-based method \cite{asadifard2010automatic} to extract the pupil from the image of the right eye, and adjust
the extracted pupil blob using morphological operations of erosion and dilation. The six steps in this process are as
follows:

\begin{enumerate}
\item Extract the right eye from the face image based on the right eye features computed as part of the face alignment step.
\item Remove all pixels that fall outside the boundaries of the polygon defined by the 6 eye features.
\item Rescale the intensity such that the 98-percentile intensity becomes 1.0 intensity and 2-percentile intensity
  becomes 0.0 intensity.
\item Define a CDF intensity threshold and convert the grayscale image to a binary image. Each pixel intensity above the
  threshold becomes 1, and otherwise becomes 0.
\item Perform an ``opening'' morphology transformation. This operation is
  useful for removing small holes in large blobs.
\item Perform a ``closing'' morphology transformation. This operation is useful for removing small objects and
  smoothing the shape of large blobs.
\end{enumerate}

The above steps have three parameters: the CDF threshold, the opening window size, the closing window size. These
parameters are dynamically optimized for each image over a discrete set of values in order to maximize the size of the
largest resulting blobs under one constraint: the largest blob must be circle-shaped (i.e. have similar height and width).

\subsubsection{Classification and Decision Pruning}\label{sec:classification}

A random forest classifier is used to generate a set of probabilities for each class from a single feature vector. The
probabilities are computed as the mean predicted class probabilities of the trees in the forest. The class probability
of a single tree is the fraction of samples of the same class in a leaf. A random forest classifier of depth 25 with an
ensemble of 2,000 trees is used for all experiments in \secref{results}. The class with the highest probability is the
one that the system assigns to the image as the ``decision''. The ratio of the highest probability to the second highest
probability is termed the ``confidence'' of the decision. A confidence of 1 is the minimum. There is no maximum. This
algorithm is used to define the \cd{Classifier} and \cd{ConfidentClassifier} algorithms with the confidence threshold
set to 1 and 10, respectively.

\subsubsection{Change and Stable State Detection}\label{sec:gaze-change-detection}

The \cd{ChangeDetector} algorithm for gaze classification uses a depth 15, 1000-tree random forest classifier based on
the following four features computed from the change between two images:

\begin{enumerate}
\item Average of dense optical flow over the bounding box of the eyes and the nose.
\item Average of dense optical flow over the bounding box of each eye.
\item Change in position of landmarks for the eyes and the nose.
\item Change in pupil position.
\end{enumerate}

The random forest classifier produces a confidence score that the image is associated with a change of state based on
the fraction of decision trees that predicted the image belongs to the ``change'' class. This score is compared with the
threshold to make the binary change detection decision. A representative image sequence and resulting change detector
output for each image is shown in \figref{change-detection}.

\subsection{Results}\label{sec:results}

\newcommand{\confusionfig}[2]{
  \begin{subfigure}[t]{0.48\textwidth}
    \includegraphics[height=2in]{images/confusion/confusion#1.png}
    \caption{#2}
  \end{subfigure}
}

\begin{figure*}
  \centering
  \confusionfig{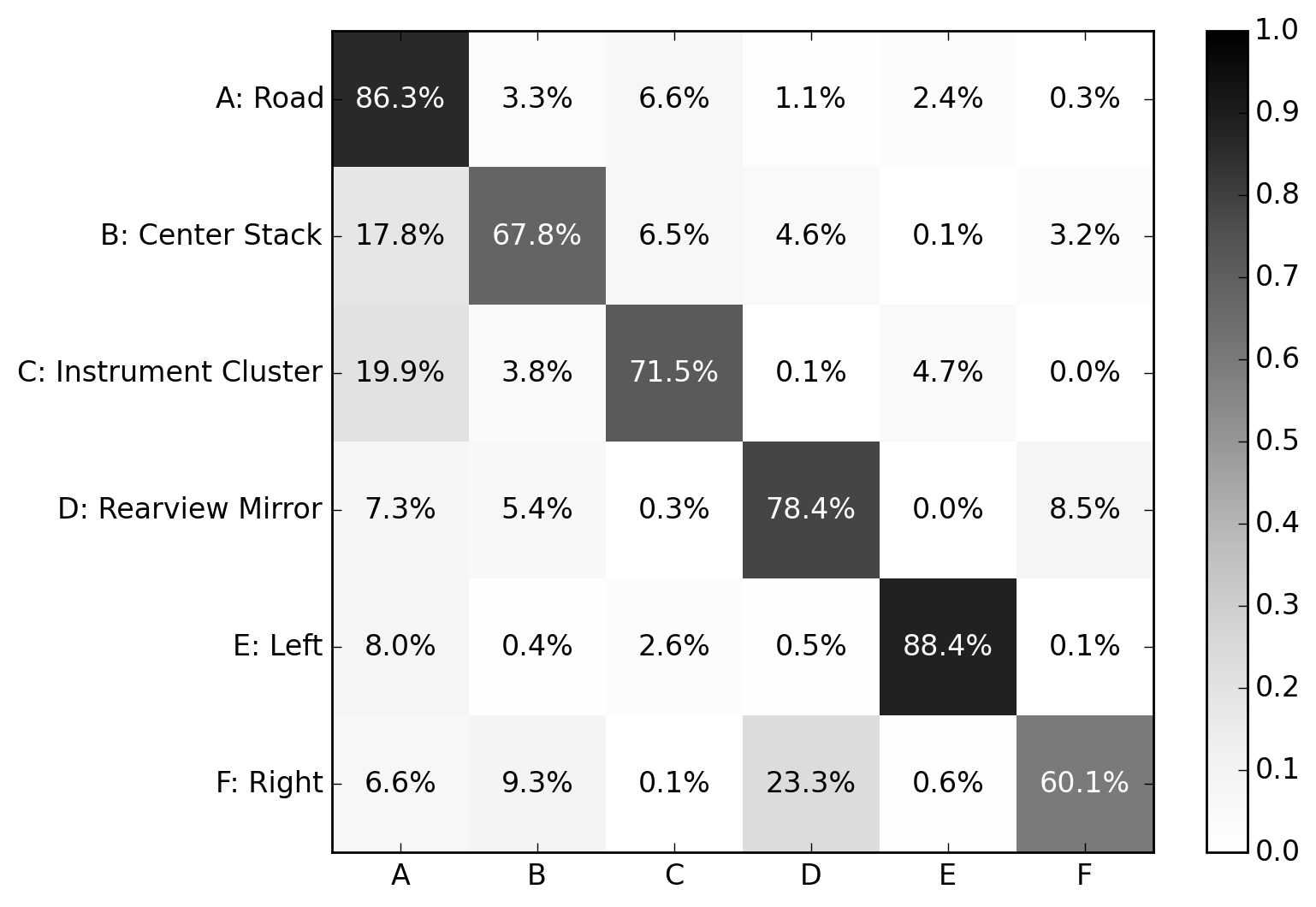}{\cd{Classifier} with average accuracy of 75.4\%.}
  \confusionfig{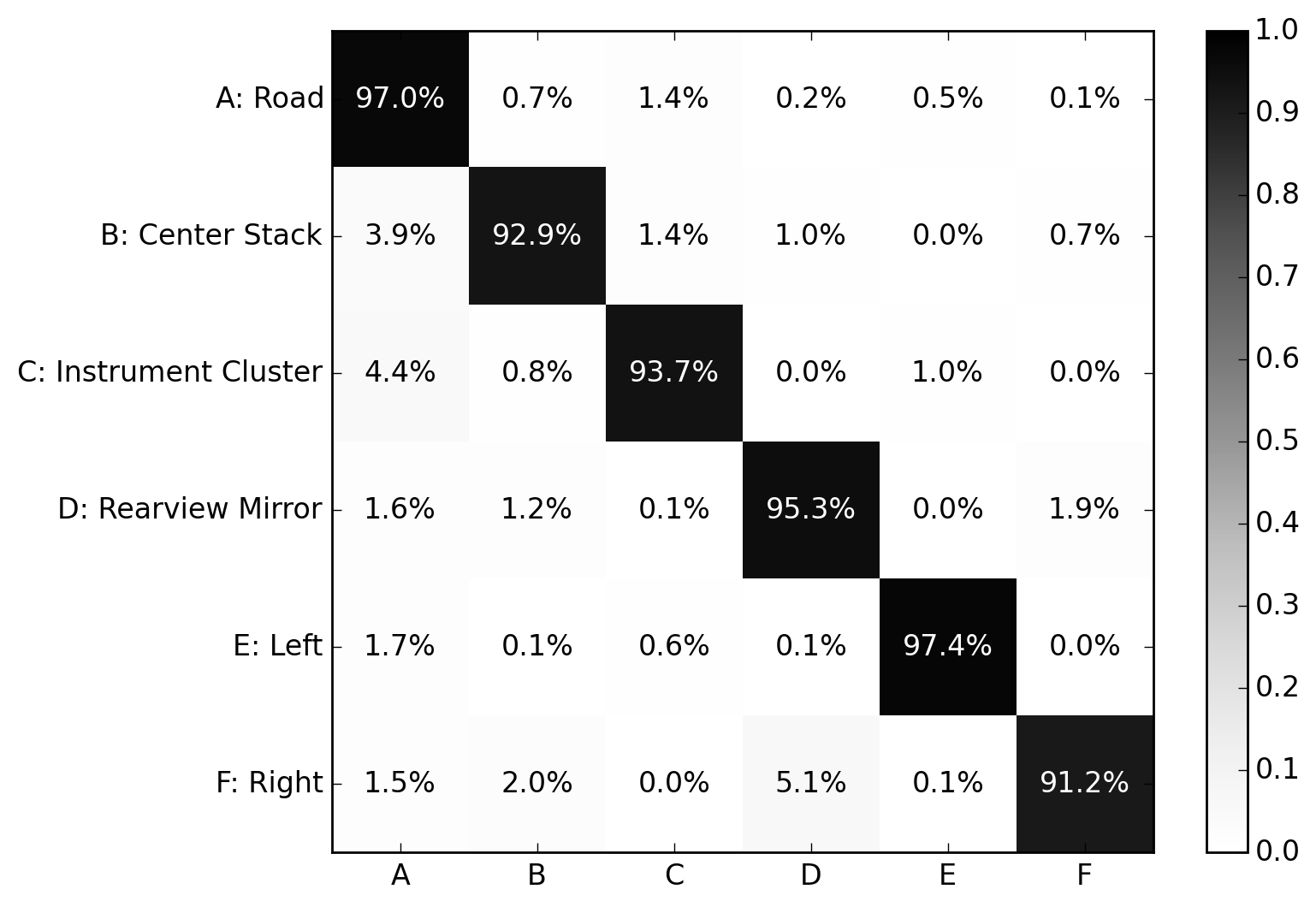}{\cd{ConfidentClassifier} with average accuracy of 94.6\%.}
  \caption{Confusion matrices for the six-region classification problem. The gaze regions are: road, center stack,
    instrument cluster, rearview mirror, left, and right.}
  \label{fig:confusion}
\end{figure*}

\begin{figure}
  \centering
  \includegraphics[width=\columnwidth]{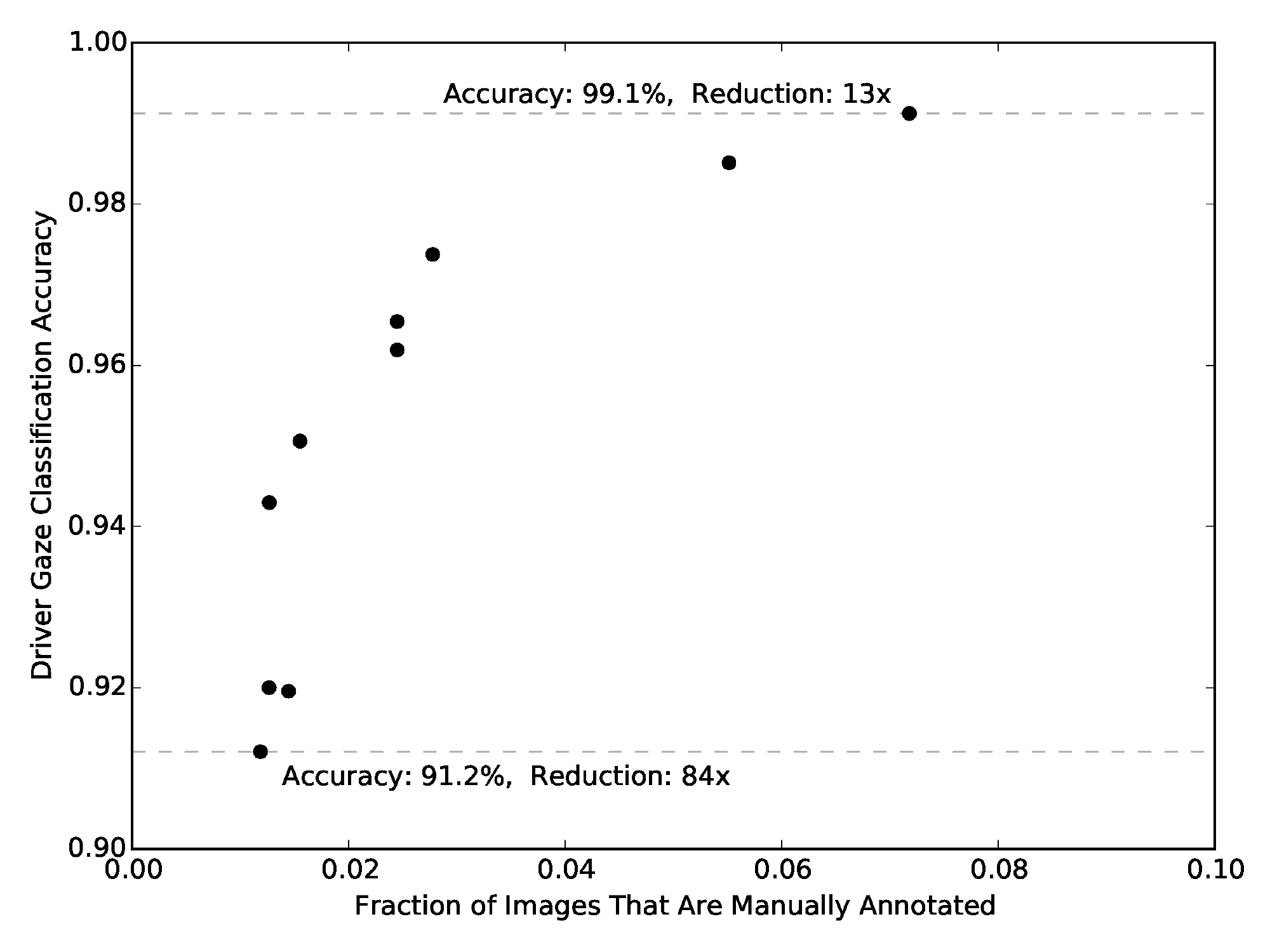}
  \caption{Parametric plot showing the empirical tradeoff between savings in human annotation effort and the accuracy of the resulting
    annotation. The biggest observed savings are an 84x reduction in human effort resulting in 91.2\% annotation
    accuracy. The highest accuracy of 99.1\% was associated with a 13x reduction in human effort.}
  \label{fig:annotation-performance}
\end{figure}

The evaluation of the gaze classification and change detection algorithms in the semi-automated annotation framework was
carried out on the 16 million video-frame dataset (see \secref{dataset}). The classifiers were ``seeded'' by assuming
that 20,000 images were first manually annotated. Then, the algorithm detailed in \secref{pipeline} was followed until
all images were annotated by human or machine. The re-training of the classifier models was performed every 20,000
manually annotated images. The confusion matrices in \figref{confusion} show the classification accuracy achieved by
\cd{Classifier} and \cd{ConfidentClassifier} algorithms given default values of parameters $\delta_{\min}$, $c_{\min}$,
and $V^u_{\min}$. The confusion matrix for \cd{Classifier} forms the conditional observation probabilities in
\eqref{viterbi} and \eqref{unchanged}.

The parametric plot in \figref{annotation-performance} shows the reduction in manual effort and annotation accuracy
achieved by varying the change detect threshold $\delta_{\min}$ from 0.1 to 0.5. The former value results in a 84x reduction
and annotation accuracy of 91.2\%. The latter value result in a 13x reduction and annotation accuracy of 99.1\%.

\section{Conclusion}\label{sec:conclusion}

We consider the problem of annotating video of an object that can be labeled as being in one of a finite set of discrete
states. The goal of the proposed solution is to significantly reduce the human effort component of the annotation. The
key insight of our approach is that we can reduce the annotation problem to a change detection problem. The performance
of the framework is evaluated on a driver gaze classification dataset composed of 16,000,000 images that were fully
annotated over 6,000 hours of direct manual annotation labor. We present a tradeoff in human effort savings and final
annotation accuracy on this dataset, showing a 13x reduction in manual annotation for an average accuracy of 99.1\% and
a 84x reduction for an average accuracy of 91.2\%.

\bibliographystyle{aaai}
\bibliography{big,lex-fridman,agelab}

\begin{thebibliography}{}

\bibitem[\protect\citeauthoryear{Asadifard and
  Shanbezadeh}{2010}]{asadifard2010automatic}
Asadifard, M., and Shanbezadeh, J.
\newblock 2010.
\newblock Automatic adaptive center of pupil detection using face detection and
  cdf analysis.
\newblock In {\em Proceedings of the International MultiConference of Engineers
  and Computer Scientists}, volume~1, ~3.

\bibitem[\protect\citeauthoryear{Bianco \bgroup et al\mbox.\egroup
  }{2015}]{bianco2015interactive}
Bianco, S.; Ciocca, G.; Napoletano, P.; and Schettini, R.
\newblock 2015.
\newblock An interactive tool for manual, semi-automatic and automatic video
  annotation.
\newblock {\em Computer Vision and Image Understanding} 131:88--99.

\bibitem[\protect\citeauthoryear{Demirkus, Clark, and
  Arbel}{2014}]{demirkus2014robust}
Demirkus, M.; Clark, J.~J.; and Arbel, T.
\newblock 2014.
\newblock Robust semi-automatic head pose labeling for real-world face video
  sequences.
\newblock {\em Multimedia Tools and Applications} 70(1):495--523.

\bibitem[\protect\citeauthoryear{Gaur and Jariwala}{2014}]{gaur2014survey}
Gaur, R.~P., and Jariwala, K.~N.
\newblock 2014.
\newblock A survey on methods and models of eye tracking, head pose and gaze
  estimation.
\newblock In {\em Journal of Emerging Technologies and Innovative Research},
  volume~1.
\newblock JETIR.

\bibitem[\protect\citeauthoryear{Ivasic-Kos, Ipsic, and
  Ribaric}{2015}]{ivasic2015knowledge}
Ivasic-Kos, M.; Ipsic, I.; and Ribaric, S.
\newblock 2015.
\newblock A knowledge-based multi-layered image annotation system.
\newblock {\em Expert Systems with Applications}.

\bibitem[\protect\citeauthoryear{Jordan and Mitchell}{2015}]{jordan2015machine}
Jordan, M., and Mitchell, T.
\newblock 2015.
\newblock Machine learning: Trends, perspectives, and prospects.
\newblock {\em Science} 349(6245):255--260.

\bibitem[\protect\citeauthoryear{Kavasidis \bgroup et al\mbox.\egroup
  }{2012}]{kavasidis2012semi}
Kavasidis, I.; Palazzo, S.; Di~Salvo, R.; Giordano, D.; and Spampinato, C.
\newblock 2012.
\newblock A semi-automatic tool for detection and tracking ground truth
  generation in videos.
\newblock In {\em Proceedings of the 1st International Workshop on Visual
  Interfaces for Ground Truth Collection in Computer Vision Applications}, ~6.
\newblock ACM.

\bibitem[\protect\citeauthoryear{Kazemi and Sullivan}{2014}]{kazemi2014one}
Kazemi, V., and Sullivan, J.
\newblock 2014.
\newblock One millisecond face alignment with an ensemble of regression trees.
\newblock In {\em Computer Vision and Pattern Recognition (CVPR), 2014 IEEE
  Conference on},  1867--1874.
\newblock IEEE.

\bibitem[\protect\citeauthoryear{King}{2009}]{dlib09}
King, D.~E.
\newblock 2009.
\newblock Dlib-ml: A machine learning toolkit.
\newblock {\em Journal of Machine Learning Research} 10:1755--1758.

\bibitem[\protect\citeauthoryear{Liu \bgroup et al\mbox.\egroup
  }{2015}]{liu2015quantitative}
Liu, Y.; Blasiak, S.; Xiao, W.; Li, Z.; and Chen, S.
\newblock 2015.
\newblock A quantitative study of video duplicate levels in youtube.
\newblock In {\em Passive and Active Measurement},  235--248.
\newblock Springer.

\bibitem[\protect\citeauthoryear{Mayer-Sch{\"o}nberger and
  Cukier}{2013}]{mayer2013big}
Mayer-Sch{\"o}nberger, V., and Cukier, K.
\newblock 2013.
\newblock {\em Big data: A revolution that will transform how we live, work,
  and think}.
\newblock Houghton Mifflin Harcourt.

\bibitem[\protect\citeauthoryear{Rabiner and
  Juang}{1986}]{rabiner1986introduction}
Rabiner, L.~R., and Juang, B.-H.
\newblock 1986.
\newblock An introduction to hidden markov models.
\newblock {\em ASSP Magazine, IEEE} 3(1):4--16.

\bibitem[\protect\citeauthoryear{Sagonas \bgroup et al\mbox.\egroup
  }{2013}]{sagonas2013300}
Sagonas, C.; Tzimiropoulos, G.; Zafeiriou, S.; and Pantic, M.
\newblock 2013.
\newblock 300 faces in-the-wild challenge: The first facial landmark
  localization challenge.
\newblock In {\em Computer Vision Workshops (ICCVW), 2013 IEEE International
  Conference on},  397--403.
\newblock IEEE.

\bibitem[\protect\citeauthoryear{Senders \bgroup et al\mbox.\egroup
  }{1967}]{senders1967attentional}
Senders, J.~W.; Kristofferson, A.; Levison, W.; Dietrich, C.; and Ward, J.
\newblock 1967.
\newblock The attentional demand of automobile driving.
\newblock {\em Highway research record} (195).

\bibitem[\protect\citeauthoryear{Shinghal and
  Toussaint}{1980}]{shinghal1980sensitivity}
Shinghal, R., and Toussaint, G.~T.
\newblock 1980.
\newblock The sensitivity of the modified viterbi algorithm to the source
  statistics.
\newblock {\em Pattern Analysis and Machine Intelligence, IEEE Transactions on}
  (2):181--185.

\bibitem[\protect\citeauthoryear{Sireesha, Vijaya, and
  Chellamma}{2013}]{sireesha2013survey}
Sireesha, M.; Vijaya, P.; and Chellamma, K.
\newblock 2013.
\newblock A survey on gaze estimation techniques.
\newblock In {\em Proceedings of International Conference on VLSI,
  Communication, Advanced Devices, Signals \& Systems and Networking
  (VCASAN-2013)},  353--361.
\newblock Springer.

\bibitem[\protect\citeauthoryear{Victor \bgroup et al\mbox.\egroup
  }{2014}]{victor2014analysis}
Victor, T.; Dozza, M.; B{\"a}rgman, J.; Boda, C.-N.; Engstr{\"o}m, J.; and
  Markkula, G.
\newblock 2014.
\newblock {\em Analysis of Naturalistic Driving Study Data: Safer Glances,
  Driver Inattention, and Crash Risk}.

\bibitem[\protect\citeauthoryear{Yuen \bgroup et al\mbox.\egroup
  }{2009}]{yuen2009labelme}
Yuen, J.; Russell, B.; Liu, C.; and Torralba, A.
\newblock 2009.
\newblock Labelme video: Building a video database with human annotations.
\newblock In {\em Computer Vision, 2009 IEEE 12th International Conference on},
   1451--1458.
\newblock IEEE.

\end{thebibliography}

\end{document}